# GujiBERT and GujiGPT: Construction of Intelligent Information Processing Foundation Language Models for Ancient Texts

Dongbo Wang[1], Chang Liu[1], Zhixiao Zhao[1], Si Shen[2], Liu Liu[1], Bin Li[3], Haotian Hu[4], Mengcheng Wu[1], Litao Lin[1], Xue Zhao[1], Xiyu Wang[1]

*1 College of Information Management, Nanjing Agricultural University, Nanjing 210095, China*

*2 Group of Science and Technology Full-text Knowledge Mining, School of Economics & Management, Nanjing University of Science and Technology, Nanjing 210094, China*

*3 College of Liberal Art, Nanjing Normal University, Nanjing 210097, China*

*4 School of Information Management, Nanjing University, Nanjing 210023, China*

**Abstract:** In the context of the rapid development of large language models, we have meticulously trained and introduced the GujiBERT and GujiGPT language models, which are foundational models specifically designed for intelligent information processing of ancient texts. These models have been trained on an extensive dataset that encompasses both simplified and traditional Chinese characters, allowing them to effectively handle various natural language processing tasks related to ancient books, including but not limited to automatic sentence segmentation, punctuation, word segmentation, part-of-speech tagging, entity recognition, and automatic translation. Notably, these models have exhibited exceptional performance across a range of validation tasks using publicly available datasets. Our research findings highlight the efficacy of employing self-supervised methods to further train the models using classical text corpora, thus enhancing their capability to tackle downstream tasks. Moreover, it is worth emphasizing that the choice of font, the scale of the corpus, and the initial model selection all exert significant influence over the ultimate experimental outcomes. To cater to the diverse text processing preferences of researchers in digital humanities and linguistics, we have developed three distinct categories comprising a total of nine model variations. We believe that by sharing these foundational language models specialized in the domain of ancient texts, we can facilitate the intelligent processing and scholarly exploration of ancient literary works and, consequently, contribute to the global dissemination of China's rich and esteemed traditional culture in this new era.

**Keywords:** Ancient book intelligent processing, large language model, domain-adaptive pre-training, digital humanities, Natural language processing

## 1 Introduction

With the rapid development of artificial intelligence, large language models are increasingly playing a key role. The combination of natural language processing comprehensive and generative language models such as BERT and GPT with datasets of vertical domains has generated numerous domainized language models. The domainized language models have greatly contributed to the development of intelligence in the corresponding domains, such as mathematics, chemistry, finance, law, and so on. As the main carrier of Chinese civilization, Chinese classics have a large span of history in time, massive data scale, and multiple categories in data types. The above characteristics of Chinese classics data have the basis for constructing a domainized Pedestal Language Model

(DPLM). Some studies have constructed the corresponding domainized base language model for Chinese classics, but the constructed model has the following shortcomings. Firstly, some models use relatively small datasets and are either simplified or traditional in terms of characters, without considering the coexistence of simplified and traditional characters. Secondly, the constructed models are mainly based on the understanding of natural language processing and lack the construction of natural language processing generative models. Finally, the determination and validation of the model performance is relatively single, not comprehensive and lacks validation on public datasets. Against this background, based on the currently available datasets of ancient books, taking into account both traditional and simplified character sets, and oriented to the two tasks of comprehension and generation of natural language processing, and with the goal of intelligent information processing of ancient books, this paper constructs the base language models for comprehension and generation of natural language processing of Chinese classics, which are named GujiBERT and GujiGPT. At present, the relevant model has been open source, see https://github.com/hsc748NLP/GujiBERT-and-GujiGPT for details.

## 2 Literature Review

### 2.1 The Pre-Trained Models of Domain Self-Adaptation

#### 2.1.1 The pre-trained model infrastructures

After the development of traditional statistical learning models and neural network models, artificial intelligence research has entered an era dominated by large-scale pre-trained models. Pre-training, fine-tuning, and transfer learning methods have become the core technologies in this field. Through self-supervised learning on large-scale unlabeled data, pre-training techniques enable models with a large number of parameters to learn richer knowledge representations and general language features. The fine-tuning process is built upon a well-pretrained base model and adapts the language model to downstream tasks by training with a small amount of carefully annotated data.
The prompt-based learning (Liu et al., 2023) has brought technological innovation to the field of natural language processing. By utilizing different templates to reshape the input format of downstream tasks, prompt-based learning bridges the gap between the task objectives during the pre-training and fine-tuning stages of large language models. This enables models to follow human instructions to accomplish various tasks, significantly enhancing the natural language processing capabilities in few-shot and zero-shot environments. Currently, most mainstream pre-training models are based on the Transformer (Vaswani et al., 2017) structure as their foundational framework. The Transformer structure is a complex network composed of multiple self-attention mechanisms, fully connected neural networks, and feed-forward neural networks stacked in a specific sequence. The Transformer structure often outperforms single-structure neural network models in various natural language understanding and generation tasks. The remarkable potential of Transformer has prompted researchers to explore expanding network depth and hidden layer dimensions to make training large-scale language models feasible. Due to the rapid improvement of hardware computing power in recent years, language models with a large number of parameters can now be updated and optimized by multiple GPUs parallel computing. A set of large pre-training models based on the Transformer structure have officially entered the stage of deep learning. Excellent variants including BERT, GPT and T5 constantly break the record of various benchmark tasks, not only bringing the researches on natural language processing into a new era but also making it possible to implement applications that were previously considered challenging tasks. In terms of the infrastructures, pre-trained models can be divided into three types: autoregressive models, autoencoder models, and encoder-decoder models. We mainly introduce the first two models, which are closely related to this research.

The currently popular GPT series models belong to the typical autoregressive models. These models usually use the decoder part of the Transformer as their basic framework. In the pre-training tasks, autoregressive models calculate the loss by predicting unidirectionally the probability of vocabulary occurrence in the text, which enables the model to possess strong text generation capabilities. The original GPT model was produced in 2018 and utilized a causal language model for pre-training. When performing downstream tasks, the GPT-1 model (Radford & Narasimhan, 2018) could retrieve the corresponding vector of the last token and input it to a fully connected layer to construct a classifier. However, this approach yielded relatively poor classification performance. The GPT-2 model (Radford et al., 2019), inheriting the infrastructure of the GPT-1 model, uses larger parameters and more data during the training process. The maximum parameter size of the GPT-2 model reaches the scale of 1.5 billion. The GPT-2 model unified all NLP tasks under the framework of text generation-related tasks. Due to the presence of translation and summarization data with indicative nature in the training corpus, it enabled the model to generate correct text in a zero-shot manner with the help of prompts. This process effectively laid the groundwork for prompt learning, which had a profound impact on subsequent research. The GPT-3 model (Brown et al., 2020) represents a significant milestone in the field. By utilizing a massively parallel GPU cluster, GPT-3 achieved a staggering parameter count of 175 billion, and the training data used reached the impressive 45 terabytes. The GPT-3 model introduced the approach of taking task examples as context input to the model. Without fine-tuning, the model which relies solely on the parameterized world knowledge and grammatical knowledge still can perform remarkably well on some tasks, approaching the performance of small-scale fine-tuned models. The recent attention-grabbing generative model is the chatGPT conversational model developed by OpenAI. It is built upon the language foundations of GPT3.5 and GPT4 (Bubeck et al., 2023). The parameter count of the latest version, GPT4, may reach the trillion-scale. Incorporating various training techniques such as code training, instruction fine-tuning, and reinforcement learning from human feedback, chatGPT integrates successfully the model's responses with natural language instructions and human values. Its remarkable comprehension and generation capabilities make it a leading model in the current stage.

The autoencoder model is another widely used model architecture. Based on the Transformer, the autoencoder model utilizes an encoder as its core component. The pre-training objective of the autoencoder model is to predict the missing content in the text using existing bidirectional information. This pre-training objective makes the autoencoder model more suitable for natural language understanding tasks. BERT(Devlin et al., 2019) is one of the most famous autoencoder models. This model utilizes a joint training loss function that combines masked language modeling and next sentence prediction tasks to learn bidirectional semantic representations of text and inter-sentence relationships. BERT has achieved remarkable success, breaking records in 11 NLP tasks. Its outstanding performance has also ushered in a new era of "pre-training + fine-tuning" in the field of artificial intelligence. RoBERTa( Liu et al., 2019) is an improved version of the BERT model. Liu et al. (2019) believed that the BERT model was not fully trained and that the next sentence prediction task lacked effectiveness. Therefore, they made several improvements to the pre-training tasks of BERT, further enhancing the model's performance. DeBERTa(He et al., 2021) is one of the most significant models recently to improve the BERT model, including proposing a disentangled attention mechanism to replace the original encoding method, using enhanced mask decoder techniques to replace BERT's masked language modeling task, and introducing adversarial training during fine-tuning to enhance the model's generalization ability on downstream tasks. Up to now, DeBERTa advances the BERT architecture model most successfully. With the rise of large-scale pre-trained models like BERT, the influence and application scope of autoencoder models without generative capabilities may be somewhat reduced. However, this does not mean that such models will be completely replaced. Due to the typically smaller parameter size and faster convergence speed of BERT-like autoencoder models compared to generative models, they are more suitable for scenarios with limited computational resources and fine-tuning on incomplete data. Additionally, autoencoder models still widely be used in some specific NLP tasks such as information retrieval and automatic recommendation.

### 2.1.2 The pre-trained models of domain knowledge enhancement

Domain self-adaptation is a transfer learning method that aims to bridge the gap between pre-training and fine-tuning data. While numerous base models have emerged in the industry, these models are typically trained in general-purpose corpora. When facing downstream tasks in specific vertical domains, the model's learned text representations during pre-training and fine-tuning may exhibit significant differences, thereby affecting the final experimental performance. Moradi et al. (2022) conducted a study in the biomedical field to compare the performance of the GPT-3 model with regular models. The research findings indicated that, in specific domains, the GPT-3 model's ability to handle small-sample learning did not surpass fine-tuned models with significantly smaller parameter sizes. This experiment highlights that without domain-specific knowledge enhancement, even the large-scale language models with the most excellent performance currently are difficult to be applied in vertical domains. To address this limitation of language models, researchers have proposed methods such as continued pre-training (Gururangan et al., 2020) for domain self-adaptation through self-supervised training on massive unlabeled corpora. This approach aims to alleviate the challenges posed by limited resources and computational constraints.

Many researchers have developed specialized models for text processing and generation in vertical domains based on BERT, RoBERTa, GPT, or other open-source large-scale language models. One such study is by Beltagy et al., (2019) , who trained the SciBERT model using computer and biomedical academic texts. They also extended the vocabulary of BERT with domain-specific scientific terms, enabling the model to generate higher-quality word embeddings for vocabulary found in academic papers. This significantly enhanced the processing capabilities of BERT-like models on academic texts. In the field of medical text processing, two small models, BioBERT (Lee et al., 2020) and ClinicalBERT (Alsentzer et al., 2019) , have been trained on biomedical text and clinical text, respectively, to address text processing problems in their respective domains. These models have shown significant improvements in performance compared to baseline models. Wang et al., (2023) trained the large model *BenTsao* (original name: HuaTuo) based on the LLaMA (Touvron et al., 2023) framework using a Chinese medical question-answering dataset. According to human evaluation, this model's answers exhibit high levels of factual accuracy and safety. In the field of law, Cui et al.,(2023) trained the large model *ChatLaw* based on two parameter levels: 13B and 33B. This model employs self-attention mechanisms to overcome potential errors in the corpus and incorporates a legal knowledge base to enhance the factuality of its answers. In the construction of specialized models for ancient Chinese text processing, researchers have recognized the language model's processing capabilities can be improved by using the pre-training methods. A relevant study, Ethan(2020), developed the GuwenBERT series of pre-training models trained on the corpus of *Dai Zhi Ge*, ancient Chinese literature, which demonstrate strong capabilities in simplified ancient Chinese text processing. However, due to the lack of a traditional Chinese word list, the model's ability to handle traditional Chinese texts is limited. KoichiYasuoka (2021) further enhanced the processing capabilities of GuwenBERT by conducting incremental training using traditional Chinese language corpora. This expansion allows the model to handle both simplified and traditional ancient Chinese texts. In our earlier research, we released two pre-training models, sikuBERT (Wang et al., 2022) and sikuGPT (Chang et al., 2023a), which were trained by using the traditional Chinese version of the electronic edition of the *Siku Quanshu* (Wen Yuan Ge edition) corpus from the publishing house. These models have achieved promising results in the comprehension and generation of traditional Chinese ancient texts.

## 2.2 Application of Pre-Trained Models in Ancient Chinese Text Processing

### 2.2.1 Ancient Chinese text understanding tasks

Ancient Chinese natural language understanding is the process of utilizing computer text processing techniques to automate and intelligent processing of ancient Chinese texts, including ancient works, books, and inscriptions. It aims to enable computers to comprehend the content of ancient Chinese texts. Ancient Chinese natural language understanding tasks mainly include automatic word segmentation, part-of-speech tagging, named entity recognition, and automatic punctuation and sentence boundary detection for ancient Chinese texts.

Ancient Chinese text automatic word segmentation and part-of-speech tagging are often treated as integrated tasks due to their close relationship. Traditional integrated tasks for ancient Chinese word segmentation and tagging often employ methods based on rules (Che et al., 2019) and ancient machine learning models (Fu et al., 2019). With the increasing number of digitized ancient textual corpora, automatic word segmentation and part-of-speech tagging techniques based on neural network models have been successfully applied to ancient Chinese text segmentation and tagging tasks, becoming the current mainstream method. Li et al., (2018) combined CNN with the sliding capsule network to extract local text features and used the routing algorithm to calculate the probability of word segmentation categories. Their approach surpassed the performance of models such as Bi-LSTM and CNN on the CTB6.0 and ACMB datasets. Hu et al.,(2022) utilized the Bi-LSTM and TextCNN models to perform automatic word segmentation and text classification tasks on Traditional Chinese Medicine (TCM) case texts. They employed the multi-task learning approach and achieved higher word segmentation accuracy than LSTM, albeit slightly lower than the pre-trained language model BERT. Tian & Guo, (2022) employed data augmentation by randomly masking input sequences and utilized the SikuBERT-BiLSTM-CRF model for automatic word segmentation and part-of-speech tagging. They achieved F-values of 94.73% and 89.19% respectively on the EvaHan2022 dataset. Feng & Li,(2023) further improved the performance of the model Tian & Guo (2022) proposed by employing a remote supervision method based on parallel corpus alignment and the SikuRoBERTa-CRF model.

The task of named entity recognition in ancient Chinese texts has received considerable attention in the context of pre-trained models. Zhou et al.,(2022) employed the Albert model to embed ancient poetry texts and utilized the BiLSTM-Attention-CRF network architecture to recognize four types of entities: time, scene, person, and location. They discovered that Albert outperformed BERT in the task. Qi et al.,(2022) observed that vocabulary in ancient Chinese texts and their corresponding modern Chinese translations exhibit similar boundary features. They proposed an adversarial transfer ancient Chinese named entity recognition model called AT-CCNER. By extracting word segmentation features from translated text using an adversarial model and transferring them to a BiLSTM-CRF model for entity recognition, the proposed approach outperformed BERT and RoBERTa models on the C-CLUE dataset. Yan et al.,(2022) proposed a pipeline-based named entity recognition (NER) approach that combines deep active learning and automatic label consolidation. The former is based on the BERT-BLSTM-CRF model, while the latter can control the quality of crowdsourcing annotation. This method outperformed all baseline models in entity recognition performance on the *Zizhi Tongjian* and *Song-Yuan Xue'an* datasets. Since ancient Chinese texts lack sentence boundary markers and punctuation (known as jùdòu 句读), automatic punctuation and sentence segmentation are crucial tasks for understanding ancient texts. Currently, pre-trained models have limited applications in this task, and neural network models remain the mainstream approach. Wang et al.,(2019) conducted automatic sentence segmentation experiments on the *Twenty-Four Histories* texts using the BiLSTM-CRF model. The results of both closed and open experiments showed that their approach outperformed models such as CRF and Bi-GRU. Han et al.,(2018) introduced radical embedding into character embedding and utilized the BiLSTM-CRF model for sequence labeling to achieve automatic sentence segmentation. Their approach demonstrated performance improvements on ancient texts in the Tang, Ming, and Qing dynasties.

Pre-trained models have been applied to some extent in the task of ancient Chinese text understanding, often surpassing the performance of machine learning and classical deep learning models in automatic word segmentation and named entity recognition tasks. However, general-purpose models like BERT and RoBERTa lack specific domain knowledge of ancient texts, which may result in insufficient extraction of lexical and syntactic features in specific conditions.

**2.2.2 Ancient text generation and cross-lingual text generation**

As another branch of ancient Chinese natural language processing, studies on ancient text natural language generation tasks are also important. Current research on ancient text generation mainly includes two subtasks: the classical poetry generation and the machine translation of ancient Chinese (i.e. cross-lingual text generation).

Due to the shorter length and strict writing formats of ancient Chinese poetry, research on automatic generation of ancient poetry started relatively early. Hu & Sun(2020) proposed an integrated approach to consolidate text formats for ancient Chinese poetry (SHI) and Song lyrics (CI). They trained a unified generation model for ancient Chinese poetry using GPT-2 and better control the poetry format by incorporating weights in the loss function. Zhao & Lee (2022) constructed an ancient poetry generation model based on Transformer-XL, which incorporates a multi-head self-attention mechanism and a segment-level recurrence mechanism. This model enables the capture of longer semantic dependencies in poetry generation. They also designed a BERT-based text fluency evaluation tool. The proposed method showed excellent performance in terms of both rhyme and fluency. Wu & Chen (2023) proposed a Word-Enhanced Transformer model. In terms of vernacular Chinese input, they employed Bi-LSTM to integrate external word segmentation features and utilized character- and word-level Transformer encoders to enhance vocabulary representation. Finally, the model generated ancient poetry using a decoder. The method achieved higher scores in BLEU and other metrics compared to Transformer, ZEN, and other generative models.

As an important step in aiding the ancient texts understanding, machine translation of ancient Chinese has also been studied widely. Jin et al.,(2022) trained a machine translation model for ancient poetry to vernacular Chinese based on Transformer. The model achieved higher scores in BLEU and human evaluation compared to GPT-2. Chang et al.,(2023b) further pre-trained GPT-2 on the Wen Yuan Ge edition of the *Siku Quanshu* (The Complete Library in Four Sections) corpus and built the SikuGPT generative model specifically for ancient Chinese text. The performance of SikuGPT outperformed models such as GPT2-chinese-ancient and Transformer in the task of translating ancient Chinese texts to vernacular Chinese in the *Twenty-Four Histories*. Shi et al., (2021) fine-tuned BERT and RoBERTa models on medical ancient texts such as *Shang Han Za Bing Lun* (Febrile Diseases) and *Jin Kui Yao Lue* (Synopsis of the Golden Chamber). They built a model for automatically generating Traditional Chinese Medicine prescriptions and validated the feasibility of their approach.

Current research on ancient text generation and cross-lingual text generation is mostly based on Seq2Seq Transformer architectures or generative Transformer decoder architectures such as GPT. These studies often involve fine-tuning or further pre-training of general pre-trained models on specific ancient texts to equip the models with the ability to generate domain-specific content. With the popularity of generative large language models like chatGPT, it is expected that the task of ancient text generation will receive more attention.

## 3 Research Methods

### 3.1 Research Process

With the continuous development of digital humanities research, language models for intelligent information processing of classics have been introduced, such as SikuBERT and SikuRoBERTa (Wang et al., 2022) for traditional ancient texts, GuwenBERT[1] for simplified ancient texts, RoBERTa-classical-Chinese-base-char[2] for both simplified and traditional ancient texts, and so on. These models continue to be trained on the basis of BERT-like models using classical texts, and show fantastic performance in the task of classic domain. Currently, the vast majority of classical language models are constructed only for traditional or simplified texts. From the perspective of classics research and protection, traditional texts or original classic texts are inevitably more valuable, while from the perspective of popularization of classical knowledge, simplified texts are more suitable. In general, learning both traditional and simplified classical corpus can simultaneously ensure the academic research value and daily value of the resulting classical model.

---

[1] https://github.com/Ethan-yt/guwenbert
[2] https://huggingface.co/KoichiYasuoka/RoBERTa-classical-Chinese-base-char

Specifically, based on the existing pre-trained models, this study trained a series of classical domain models using traditional, simplified, and mixed traditional and simplified classical texts, respectively. On this basis, the models trained in this study and the related baseline models were fine-tuned with relevant datasets to adapt the models to the downstream tasks in the field of ancient books, and then the performance of the models was verified.

### 3.2 Data and Models

The data used in this model training comes from the website of Daizhige[3], which contains about 1.7 billion words of classical texts, and divides them into ten categories, such as Confucianism, Buddhism, History, and so on, with some of them in simplified Chinese characters and the others in traditional Chinese characters. In this study, the traditional and simplified data in the Daizhige corpus were firstly converted into traditional and simplified Chinese characters respectively, and this process was carried out by using the Chinese Character Simplified-Traditional Conversion Text Intelligence System[4] developed by Xiamen University, which provides three forms of online conversion, Word plug-in, and stand-alone software to realize the conversion of traditional and simplified Chinese characters, and in view of the large scale of the data, the data were converted into traditional and simplified Chinese characters by using the stand-alone software in the present study. Secondly, the converted data were cleaned to remove special symbols and invalid characters. Finally, the traditional and simplified data of Daizhige corpus are obtained respectively, and the two data are merged to form the mixed data of traditional and simplified, and the model is trained using the traditional, simplified and mixed data respectively.

In terms of baseline model selection, different baseline models are chosen according to the training corpus and pre-training tasks. Specifically, SikuBERT, SikuRoBERTa, and SikuGPT are used as baseline models for the traditional Chinese classics corpus and the mixed traditional and simple classics corpus, respectively, which are all trained by our group based on the traditional Chinese corpus of the Siku Quanshu. For the Simplified Chinese classics corpus, BERT-base-Chinese[5], Chinese-RoBERTa-wwm-ext[6] and GPT2-Chinese-cluecorpussmall[7] are used as baseline models.

### 3.3 Pre-trained Methods

The BERT model, as a typical self-encoding language model, has a bi-directional encoder as its modeling framework, and the pre-training task used is the Masked Language Model (MLM). Masked Language Model (MLM) is a pre-training task for training bi-directional text representation, in which 15% of the vocabulary is randomly masked and parameter updating is accomplished by predicting the masked tokens. GPT, as a typical autoregressive language model, with a unidirectional decoder model framework, employs the training method of Causal Language Model (CLM). Causal language modeling is a pre-training task for training unidirectional text representations, in which the model only needs to predict the words that appear next based on the words on one side of the input sentence, and then use the cross-entropy loss function to update the parameters of the model. In contrast to masked language models, causal language models allow predictions to be made with reference to only one side of the content. When the training goal is is to learn a good representation of the input text, the Masked Language Model (MLM) is undoubtedly a better choice due to its ability to consider the context simultaneously; and when the training goal is to generate fluent text, the unidirectional causal language model is similar to the way human beings write, which can better enhance the model's creative ability. In summary, in this study, the Masked Language Model (MLM) is used to train the BERT and RoBERTa models, and the Causal Language Model (CLM) is used to train the GPT model, and the training process is based on the Transformers framework provided by Huggingface.

---

[3] http://daizhige.org/
[4] http://jf.xmu.edu.cn/
[5] https://huggingface.co/bert-base-chinese
[6] https://huggingface.co/hfl/chinese-roberta-wwm-ext
[7] https://huggingface.co/uer/GPT2-chinese-cluecorpussmall

### 3.4 Downstream Task

As mentioned earlier, BERT-like models are more suitable for dealing with language comprehension tasks, while GPT-like models are more suitable for dealing with language generation tasks. Therefore, the test for model performance is also divided into two parts: natural language understanding evaluation task and natural language generation evaluation task. Specifically, the natural language understanding evaluation task includes four evaluation tasks: named entity recognition, lexical integration labeling, sentence breaking and punctuation of classical texts, while the natural language generation evaluation task is the evaluation of machine translation of classical texts. In the evaluation process, since the datasets used are all traditional corpus, it is also necessary to convert the corpus into simple and traditional and merge the traditional and simplified data in order to evaluate the models trained on different corpus respectively. We use the simplified corpus to evaluate GujiBERT_jian, GujiRoBERTa_jian, BERT-base-Chinese, Chinese-RoBERTa-wwm-ext, and GuwenBERT for language comprehension, and GujiGPT_jian, GPT2-Chinese- ancient for language generation. ancient for language generation task evaluation. The traditional corpus was used to evaluate GujiBERT_fan, GujiRoBERTa_fan, SikuBERT, SikuRoBERTa, BERT-base-Chinese, and Chinese-RoBERTa-wwm-ext for the traditional classics text comprehension task, and GujiGPT_fan, SikuGPT, and GPT2-base-Chinese for the traditional ancient text generation task evaluation. Evaluating GujiBERT_jian_fan, GujiRoBERTa_jian_fan, BERT-base-Chinese, Chinese-RoBERTa-wwm-ext, and RoBERTa-classical-Chinese-base-char for the traditional-simplified ancient text comprehension task using the traditional-simplified mixed corpus. Mixed classical text comprehension task was assessed, and GujiGPT_jian_fan was assessed for the traditional-simplified mixed classical text generation task.

## 4 Results

### 4.1 Pre-trained Models

After the statistical extraction of the Daizhige corpus, all the Chinese characters that can be displayed in utf-8 encoding are obtained, and the word list is expanded on the bas is of the word list of the baseline model, and the expansion of the word list is shown in Table 1. After that, the cleaned Daizhige corpus is divided into training set and validation set according to the ratio of 99:1, and the model continues to be trained based on the transformers framework using MLM and CLM methods, and the model training parameters are shown in Table 2.

Table 1 Vocab expansion

| pre-trained models | number of words expanded |
|---|---|
| SikuBERT | 4918 |
| SikuRoBERTa | 4918 |
| SikuGPT | 4261 |
| BERT-base-Chinese | 13581 |
| Chinese-RoBERTa-wwm-ext | 13581 |
| GPT2-Chinese-cluecorpussmall | 4272 |

Table 2 Training parameter settings

| parameters | values |
|---|---|
| learning_rate | 2e-5 |
| block_size | 512(BERT)/1024(GPT) |
| processing_num_workers | 16 |
| num_train_epochs | 5 |
| per_device_train_bach_size | 32(BERT)/16(GPT) |
| per_device_eval_batch_size | 64(BERT)/32(GPT) |

For the initial validation of the model performance, perplexity is chosen as a metric. Perplexity is a metric for evaluating language performance based on the probability of sentences in the test set, and the basic idea is that a language model that assigns higher probability values to the sentences in the test set is better, and when the language model has been trained, and the sentences in the test set are all normal sentences, the trained model is the one that has higher probability in the test set with higher probability is better. In perplexity calculation, a sentence can be represented as:

$$S = W_1, W_2, …, W_n \tag{1}$$

The probability that a sentence occurs is:

$$P(S) = P(W_1, W_2, …, W_n) = P(W_1)P(W_2|W_1) … P(W_n|W_1, W_2, …, W_{n-1}) \tag{2}$$

Based on this, the perplexity is calculated as:

$$PPL = P(w_1 w_2 … w_n)^{-\frac{1}{n}} = \sqrt[n]{\frac{1}{P(w_1 w_2 … w_n)}} \tag{3}$$

The perplexity of the pre-trained models and their corresponding baseline models are shown in Table 3, and it can be seen that the GPT-based models have higher perplexity compared to the BERT-based models, while the RoBERTa-based models have slightly lower perplexity than the BERT-based models. Meanwhile, the perplexity of the models trained based on the Siku series models is lower than that of the models trained based on the Simplified Chinese models, which also shows the superiority of the Siku series models in the field of ancient text specialization. Overall, compared with the baseline model, the perplexity of the Guji series models, especially the models obtained from the GPT class model and the mixed traditional and simplified Chinese corpus training, decreases significantly, which indicates that the pre-trained models learn the linguistic knowledge in the corpus to a certain extent during the model training process.

Tabel 3 Perplexity of pre-trained models

| models | perplexity | base-models | perplexity |
|---|---|---|---|
| GujiBERT_fan | 11.0313 | SikuBERT | 69.5678 |
| GujiRoBERTa_fan | 9.4185 | SikuRoBERTa | 79.5975 |
| GujiGPT_fan | 34.2315 | SikuGPT | 154.985 |
| GujiBERT_jian | 12.3738 | BERT-base-Chinese | 74.0597 |
| GujiRoBERTa_jian | 11.7319 | Chinese-RoBERTa-wwm-ext | 85.1540 |
| GujiGPT_jian | 42.5118 | GPT2-Chinese-cluecorpussmall | 910.0321 |
| GujiBERT_jian_fan | 9.5282 | SikuBERT | 188.3149 |
| GujiRoBERTa_jian_fan | 9.3962 | SikuRoBERTa | 253.3989 |
| GujiGPT_jian_fan | 33.7822 | SikuGPT | 448.0125 |

## 4.2 Natural Language Understanding Evaluation Task

The natural language understanding tasks in this study are all sequence annotation tasks, and according to the convention, Precision (P), Recall (R), and Functional Mean (F1) are used as rubrics (Atterer & Schütze, 2007). The original corpora used in this phase are all traditional ancient Chinese corpora, and in order to fit the applicable dimensions of the various models, we also consider adding simplified corpora and mixed corpora for the experiments. In the following, all the experiments conducted on Simplified Chinese are based on Simplified Chinese texts converted with the OpenCC library's Simplified-Traditional conversion function, while the Simplified+Traditional experiments are based on a mixed corpus, which is twice as large as the other experiments.

### 4.2.1 Named Entity Recognition

Through the recognition of words with specific meanings, such as human-named entities, place-named entities and time-word entities in ancient books, a finer-grained slicing of the ancient classics can be realized, which in turn helps the construction of the knowledge graph of classical texts and the application of classical text retrieval and translation.

In this study, we utilize the named entity recognition corpus of the Shiji to fine-tune the training of each base model, and on the basis of this, we realize the model construction and testing for the named entity recognition task. The Shiji is a detailed narrative, in which the characters, places and times recorded have great research value for named entity recognition in the field of ancient books. In this study, the corpus is divided into training set and test set according to the ratio of 9:1. The test results are shown in Table 4, from which it can be seen that the "Siku" and "Guji" series models developed by our group are better in the task of recognizing named entities in traditional Chinese ancient texts, with the F1 value about 4% higher than that of the baseline model RoBERTa; and in the task of sentence breaks in simplified Chinese ancient texts, the F1 value of Guji is about 4% higher than that of the baseline model RoBERTa. On the task of sentence breaking in simplified Chinese, GuwenBERT achieves the best performance, with an F1 value of 93.73%; on the mixed corpus of traditional Chinese and simplified Chinese, GujiBERT_jian_fan achieves the best performance, with an F1 value of 97.02%.

Table 4 Results of Named Entity Identification

| training corpus | models | P(%) | R(%) | F1(%) |
|---|---|---|---|---|
| traditional | GujiBERT_fan | 92.8 | 94.28 | 93.53 |

| training corpus | models | P | R | F1 |
|---|---|---|---|---|
| traditional | GujiRoBERTa_fan | 92.71 | 93.92 | 93.3 |
| traditional | SikuBERT | 92.5 | 93.92 | 93.2 |
| traditional | SikuRoBERTa | 92.43 | 93.73 | 93.07 |
| traditional | BERT-base-Chinese | 90.16 | 92.12 | 91.12 |
| traditional | Chinese-RoBERTa-wwm-ext | 88.86 | 90.65 | 89.75 |
| simplified | GujiBERT_jian | 91.93 | 93.55 | 92.73 |
| simplified | GujiRoBERTa_jian | 91.7 | 93.39 | 92.53 |
| simplified | BERT-base-Chinese | 87.36 | 89.82 | 88.57 |
| simplified | Chinese-RoBERTa-wwm-ext | 88.37 | 90.39 | 89.37 |
| simplified | GuwenBERT | 92.84 | 94.66 | 93.73 |
| simplified + traditional | GujiBERT_jian_fan | 97.65 | 97.29 | 97.02 |
| simplified + traditional | GujiRoBERTa_jian_fan | 95.91 | 96.78 | 96.34 |
| simplified + traditional | BERT-base-Chinese | 95.75 | 96.53 | 96.14 |
| simplified + traditional | Chinese-RoBERTa-wwm-ext | 94 | 95.33 | 94.66 |
| simplified + traditional | RoBERTa-classical-Chinese-base-char | 94.76 | 94.83 | 94.76 |

### 4.2.2 Segmentation and Lexical Labeling

In the process of constructing the base language model for intelligent information processing of ancient books, the quality of participle and lexical annotation has a key impact on text mining, knowledge graph presentation, and named entity recognition of classics. In this study, based on the corpus of Zuozhuan, the base model is fine-tuned, trained, and tested in order to construct an integrated labeling model of ancient Chinese participle and lexical annotation. The corpus of Zuozhuan is created and released by the Chinese information processing research team led by Prof. Chen of Nanjing Normal University, which consists of about 180,000 words, uses 17 kinds of classical markers designed by themselves, and has been proofread for four times, so that the quality of the corpus is guaranteed. In this study, the database is divided into three types of datasets: traditional Chinese characters, simplified Chinese characters, and mixed simplified and traditional Chinese characters, to explore the performance of each base model in the three types of datasets, and the test results are shown in Table 5.

Table 5 Results of Segmentation and Lexical Labeling

| training corpus | models | P(%) | R(%) | F1(%) |
|---|---|---|---|---|
| traditional | GujiBERT_fan | 90.49 | 91.03 | 90.74 |
| traditional | GujiRoBERTa_fan | 90.47 | 91.02 | 90.73 |
| traditional | SikuBERT | 90.48 | 90.98 | 90.68 |
| traditional | SikuRoBERTa | 90.21 | 90.87 | 90.52 |
| traditional | Bert-base-Chinese | 89.62 | 90.36 | 89.93 |
| traditional | Chinese-RoBERTa-wwm-ext | 88.96 | 89.95 | 89.43 |
| simplified | GujiBERT_jian | 90.66 | 91.25 | 90.94 |
| simplified | GujiRoBERTa_jian | 90.04 | 90.72 | 90.36 |
| simplified | Bert-base-Chinese | 89.75 | 90.49 | 90.07 |
| simplified | Chinese-RoBERTa-wwm-ext | 89.17 | 90.07 | 89.6 |
| simplified | GuwenBERT | 90.56 | 91.27 | 90.89 |
| simplified + traditional | GujiBERT_jian_fan | 93.76 | 93.83 | 93.76 |
| simplified + traditional | GujiRoBERTa_jian_fan | 93.22 | 93.48 | 93.3 |
| simplified + traditional | Bert-base-Chinese | 93.05 | 93.36 | 93.16 |
| simplified + traditional | Chinese-RoBERTa-wwm-ext | 92.37 | 92.76 | 92.54 |
| simplified + traditional | RoBERTa-classical-Chinese-base-char | 92.56 | 92.64 | 92.58 |

As can be seen from Table 5, in the traditional split word lexical integration annotation task, GujiBERT_fan has the best performance, with 90.49%, 91.03%, and 90.74% precision (P), recall (R), and F1 score, respectively, which is a big improvement from the baseline model. Chinese-RoBERTa-wwm-ext has the worst performance, with the lowest accuracy, recall, and F1. In the Simplified Segmentation Lexical Integration task, GujiBERT_jian has the highest F1 score, reconciliation mean and shows better performance in segmentation lexical annotation, but the recall is lower than that of GuwenBERT by 0.02 percentage points. GujiBERT_jian_fan has the best performance in the task of mixed split word lexical integration of Simplified Chinese and Traditional Chinese, and the model is also the best tested among all base models, obtaining the highest accuracy, recall, and tonal mean. RoBERTa has the lowest accuracy and tonal mean, and RoBERTa-classical-Chinese-base-char has the lowest recall. By testing the three types of datasets, it can be found that Bert-based base models, such as GujiBERT_fan, GujiBERT_jian, GuwenBERT, and GujiBERT_jian_fan, may be better at the task of integrating the lexical annotation of ancient Chinese participles.

### 4.2.3 Sentences Breaking

The texts recorded in the original classics usually do not have punctuation marks. Sentence breaking aims at adding sentence separators to unpunctuated ancient texts, which not only enhances the readability of the ancient Chinese texts, but also guarantees the processing of different levels of textual information.

In this study, we utilize the corpus of the Twenty-Four Histories to fine-tune the models, construct an ancient Chinese sentence breaking model for the automatic sentence breaking task, and test the performance of the model. The ancient Chinese corpus of the Twenty-Four Histories selected in this study is extracted from The Complete Translation of the Twenty-Four Histories compiled under the auspices of Xu Jialu and others, which has been carefully proofread by many experts and is of high quality, and the corpus contains sentence breakage markers. In this study, 300,000 sentences are selected, and the corpus is divided into training set and test set according to the ratio of 9:1, and the test results are shown in Table 6. From the data in Table 6, it can be seen that the "Siku" and "Guji" series of models developed by our group have obvious advantages over the general models in the task of sentence breaking in traditional Chinese. The best performance is achieved by GujiRoBERTa_fan, with an F1 value of 86.31%; in the task of simplified ancient text sentence breakage, GuwenBERT achieves the best performance, and GujiRoBERTa_jian achieves the second best performance, with a difference of about 3 percentage points. This is due to the fact that GuwenBERT's base model is Chinese-BERT-wwm for the Simplified Chinese task, and the corpus on which GuwenBERT continues to be pre-trained is a large-scale Simplified Ancient Chinese, whereas GujiBERT was obtained based on SikuBERT for Traditional Ancient Chinese, and SikuBERT's stronger Traditional Ancient Chinese domain may weaken the GujiBERT's ability to deal with Simplified Ancient Chinese; on the mixed corpus of Traditional Ancient Chinese and Simplified Ancient Chinese, RoBERTa-classical-Chinese-base-char performs optimally, which is on the one hand, due to the fact that the base model of RoBERTa-classical-Chinese-base-char is GuwenBERT, on the other hand, although the Simplified Chinese and Traditional Chinese corpus in this task have the same number of characters, the Traditional Chinese has more characters with the same morphology as its Simplified form, so the total number of Simplified Chinese characters is more, which makes RoBERTa-classical-Chinese-base-char perform better. The above results also corroborate the effectiveness of continued pre-training of the model's domainization, highlighting the need for pre-training language models for traditional Chinese.

Table 6 Results of Sentences Breaking

| training corpus | models | P(%) | R(%) | F1(%) |
|---|---|---|---|---|
| traditional | GujiBERT_fan | 84.90 | 85.85 | 85.37 |
| traditional | GujiRoBERTa_fan | 85.87 | 86.75 | 86.31 |
| traditional | SikuBERT | 84.12 | 85.16 | 84.63 |
| traditional | SikuRoBERTa | 85.06 | 85.96 | 85.51 |
| traditional | BERT-base-Chinese | 78.78 | 79.92 | 79.35 |
| traditional | Chinese-RoBERTa-wwm-ext | 77.34 | 78.16 | 77.75 |
| simplified | GujiBERT_jian | 83.29 | 84.20 | 83.74 |
| simplified | GujiRoBERTa_jian | 83.47 | 84.38 | 83.92 |
| simplified | BERT-base-Chinese | 78.86 | 79.74 | 79.30 |
| simplified | Chinese-RoBERTa-wwm-ext | 77.44 | 78.15 | 77.80 |
| simplified | GuwenBERT | 86.96 | 87.41 | 87.18 |
| simplified + traditional | GujiBERT_jian_fan | 86.03 | 86.88 | 86.45 |
| simplified + traditional | GujiRoBERTa_jian_fan | 86.44 | 87.12 | 86.78 |
| simplified + traditional | BERT-base-Chinese | 80.90 | 81.76 | 81.33 |
| simplified + traditional | Chinese-RoBERTa-wwm-ext | 79.98 | 80.61 | 80.29 |
| simplified + traditional | RoBERTa-classical-Chinese-base-char | 87.65 | 88.04 | 87.85 |

### 4.2.4 Automatic Punctuation

Automatic punctuation of ancient Chinese refers to the use of Chinese current "Punctuation Usage" (GB/T 15834-2011) specification for the original ancient texts without punctuation automatically add punctuation, not only makes the processed ancient texts more in line with the reading habits of modern people, but also is the basis for the realization of the fine-grained processing of ancient texts. In this section, we use the twenty-five pre-Qin canonical texts that have been manually labeled and proofread by our group for automatic punctuation experiments. In order to fully improve the generalization ability of the model, this study selects 189798 corpora with sufficient length for training to ensure that the model can learn richer semantic representations. In the data processing stage, this study divides the dataset according to the ratio of 8:1:1, and after removing all the punctuation in the dataset, the data are processed using the BIO annotation format to ensure that the label corresponding to the previous character at the punctuation position in the original text is the punctuation label in the label set. For the sample "The disciple asked, "The husband saw Laozi." the specific processing results are shown in the Table 7.

Table 7 Examples of annotated data

| texts | labels |
|---|---|
| 弟 | O |
| 子 | O |
| 問 | O |
| 曰 | B-f |
| 夫 | O |
| 子 | O |
| 見 | O |
| 老 | O |
| 聃 | B-d |

Table 8 shows the performance of various models on the task of automatic punctuation of Chinese classics. On the task of automatic punctuation of Chinese classics, all the ancient Chinese base models have gained advantages over the generalized models, among which, on the task of punctuation of traditional ancient Chinese, the performance of GujiBERT_fan and GujiRoBERTa_fan models we developed is significantly better than that of the "Siku" series models. Among them, GujiRoBERTa_fan achieves the best performance with the highest F1 value of 76.00%. On the task of automatic punctuation of simplified text, GuwenBERT achieves the best performance with the highest F1 value of 77.19%, which is slightly higher than that of "Guji" series model, and this difference may be related to the word list and the training corpus. The GujiRoBERTa_jian_fan model achieves the best performance on the mixed corpus of traditional Chinese and simplified Chinese, with the highest F1 value of 78.53%, and the RoBERTa-classical-Chinese-base-char model achieves the second best performance, with the highest F1 value of 78.47%, due to the fact that it is trained from the GuwenBERT model. GuwenBERT model, it also inherits the efficient automatic punctuation of GuwenBERT model. Taking the above tasks together, we find that the GuwenBERT family of models may be more suitable for handling automatic sentence breaking and punctuation tasks, probably because GuwenBERT streamlines the vocabulary lists and retrains the word embedding representations.

Table 8 Results of Punctuation

| training corpus | models | P(%) | R(%) | F1(%) |
|---|---|---|---|---|
| traditional | GujiBERT_fan | 75.96 | 76.93 | 76.00 |
| traditional | GujiRoBERTa_fan | 76.06 | 77.03 | 76.09 |
| traditional | SikuBERT | 75.68 | 76.60 | 75.69 |
| traditional | SikuRoBERTa | 75.73 | 76.97 | 75.91 |
| traditional | BERT-base-Chinese | 73.99 | 74.85 | 73.91 |
| traditional | Chinese-RoBERTa-wwm-ext | 73.94 | 74.59 | 73.71 |
| simplified | GujiBERT_jian | 75.76 | 76.51 | 75.68 |
| simplified | GujiRoBERTa_jian | 75.35 | 76.30 | 75.42 |
| simplified | BERT-base-Chinese | 74.50 | 75.07 | 74.20 |
| simplified | Chinese-RoBERTa-wwm-ext | 74.47 | 75.00 | 74.17 |
| simplified | GuwenBERT | 77.02 | 78.45 | 77.19 |
| simplified + traditional | GujiBERT_jian_fan | 77.53 | 78.32 | 77.92 |
| simplified + traditional | GujiRoBERTa_jian_fan | 78.14 | 78.93 | 78.53 |
| simplified + traditional | BERT-base-Chinese | 75.69 | 75.15 | 75.42 |
| simplified + traditional | Chinese-RoBERTa-wwm-ext | 75.93 | 75.84 | 75.88 |
| simplified + traditional | RoBERTa-classical-Chinese-base-char | 78.06 | 78.89 | 78.47 |

## 4.3 Natural Language Generation Evaluation Task

Natural Language Generation (NLG) task is an important branch in the field of intelligent information processing, covering subtasks such as text summarization, machine translation, dialog systems and so on. In this study, in order to investigate the specific performance of base models on the natural language generation task, the models are evaluated using the machine translation subtask. In this evaluation task, the book "Twenty-four Histories in Full Translation", a key book of the National "Tenth Five-Year" Publishing Plan, was chosen as the data source. One million sets of roughly processed parallel corpus of the Twenty-four Histories were obtained through the Aligner tool and several rounds of manual proofreading. In order to improve the accuracy of the task evaluation, this study further deletes the parallel corpus by calculating the similarity between the original and the translated texts, retains the aligned corpus whose similarity between ancient Chinese and modern Chinese is between 0.85 and 0.98, totaling about 300,000 groups, and uses the aligned corpus for the construction of the final machine translation model, with the specific forms of the corpus input shown in Table 9.

Table 9 Examples of annotated data

| ID | parallel corpus |
|---|---|
| 1 | {"acn": "累遷尚書二千石郎中。", "mcn": "屢經升遷爲尚書二千石郎中。"} |
| 2 | {"acn": "賈繼春，新鄉人。", " mcn ": "賈繼春是新鄉人。"} |
| 3 | {" acn ": "永淳二年析建昌復置。", " mcn ": "永淳二年分割建昌縣重設。"} |

On the basis of completing the corpus construction, in order to accurately assess the actual translation ability (text generation ability) from ancient Chinese to modern Chinese of each model, the study adopted BLEU as the main evaluation index of the experiment. All models used the same experimental configuration information, and after five rounds of training, the evaluation results of the generated content of each model are shown in Table 9.

Table 9 Results of Machine Translation

| training corpus | models | BLEU1 | BLEU2 | BLEU3 | BLEU4 |
|---|---|---|---|---|---|
| traditional | SikuGPT | 0.714 | 0.505 | 0.383 | 0.297 |
| traditional | GujiGPT_fan | 0.715 | 0.507 | 0.385 | 0.299 |
| traditional | GPT2-base-Chinese | 0.715 | 0.507 | 0.385 | 0.299 |
| simplified | GPT2-Chinese-ancient | 0.712 | 0.503 | 0.381 | 0.295 |
| simplified | GujiGPT_jian | 0.715 | 0.505 | 0.383 | 0.298 |
| simplified + traditional | GujiGPT_jian_fan | 0.715 | 0.507 | 0.385 | 0.299 |

On the one hand, the performance of the language models trained based on the Simplified, Traditional and Combined corpus of GujiGPT all reach 0.715 on the BLEU1 metric, which indicates that the translation results of this kind of models have a high accuracy. On the other hand, both models GujiGPT_fan and GujiGPT_jian_fan also reach 0.299 on the BLEU4 metric, which indicates that the model translation results have good fluency. This indicates that the natural text generation model based on GujiGPT corpus training proposed in this study has better text generation ability and achieves the best evaluation results in this experimental machine translation task.

## 5 Research Conclusion and Outlook

Based on summarizing the current research on intelligent processing of ancient books, this article proposes the necessity of constructing dedicated models for processing ancient texts that are suitable for different dimensions. Three categories and nine types of models were designed according to user processing preferences, including models for processing traditional Chinese ancient books, simplified Chinese ancient books, and mixed-style ancient books, achieving full coverage of intelligent processing tasks for ancient books under different fonts. In our trained models, the performance of three models, GujiBERT_fan, GujiRoBERTa_fan, and GujiBERT_jian, all surpassed the original baseline model, indicating that increasing the scale of training corpora of the same font is beneficial for improving model capabilities. In the simplified Chinese models, our GujiBERT_jian and GujiRoBERTa_jian models also outperformed the baseline model but showed slightly weaker performance than GuwenBERT in some tasks, making them alternative choices. In the corpus of mixed traditional and simplified Chinese, our GujiBERT_jian_fan and GujiRoBERTa_jian_fan achieved the best results in Named Entity Recognition (NER), Part-of-Speech Tagging, and Punctuation tasks. However, their sentence segmentation performance was slightly lower than the RoBERTa-classical-Chinese-base-char model trained with GuwenBERT. In generative tasks, Chinese models trained on traditional Chinese corpora performed well, but there was no significant difference overall. This may indicate that small-scale generative models are less sensitive to variations in corpora compared to natural language understanding models.

The research results of this article demonstrate that continuing to train models using ancient Chinese corpora in a self-supervised manner can effectively enhance the model's ability to handle downstream tasks. The use of training data in different fonts, adjusting the amount of pre-training data, and selecting different baseline models all have an impact on the final experimental results. However, overall, there is no significant difference in performance among various models trained using ancient Chinese data, and the final performance of executing some tasks is comparable. Since the total volume of ancient books is no longer increasing, under the same parameter size and model structure, models pre-trained on Classical Chinese corpora are more likely to approach their upper limits, and the benefits of training such models are diminishing. Therefore, the construction of models for Classical Chinese processing also needs to seek new breakthroughs. Specifically, in future related research, our team will make full use of our computational and data advantages to explore the methods of building a large-scale infrastructure for Classical Chinese models using generative large language models and multimodal pre-training techniques, continuing to contribute to intelligent processing of ancient books, digital humanities research, and the promotion of Chinese excellent traditional culture.

## Acknowledgement

The authors acknowledge the National Social Science Foundation of China (Grant Numbers: (21&ZD331)) for financial support. We thank all the volunteers and all publications support and staff who wrote and provided helpful comments on previous versions of this document.

## Conflict of interest statement

The authors declared that they have no conflicts of interest to this work.